\documentclass[runningheads]{llncs}
\usepackage[T1]{fontenc}
\usepackage{graphicx}
\usepackage{makecell}
\usepackage{url}
\usepackage{subcaption}
\usepackage{hyperref}
\def\argmin{\mathop{\rm argmin}\limits}

\usepackage{xcolor}
\usepackage{newunicodechar}
\usepackage{mathrsfs}
\usepackage{algorithm}
\usepackage{algpseudocode}
\usepackage{amsmath,amssymb,amsfonts}
\usepackage{comment}
\newunicodechar{≈}{\approx}
\newcommand{\tb}{\textbf}

\usepackage{color}

\begin{document}
\title{LiFT: Local Search via Linear Programming for Overfitting-Controlled Transformers}
\titlerunning{LiFT: LP-Based Fine-Tuning for Transformers}

\author{Abhishek Shukla\inst{1,4}\orcidID{0000-0002-0402-2982} \and Anikeit Khanna\inst{2}\orcidID{0009-0009-0442-0583} \and Ankur Sinha\inst{3,4}\orcidID{0000-0002-0464-6736} \and Faiz Hamid\inst{1}\orcidID{0000-0002-3058-0152}}
\authorrunning{A. Shukla et al.}
\institute{
Department of Management Sciences,
Indian Institute of Technology Kanpur,\\
Kanpur-208016, Uttar Pradesh, India\\
\email{\{abhiskl,fhamid\}@iitk.ac.in}
\and
Department of Civil Engineering,
Indian Institute of Technology Kanpur,\\
Kanpur-208016, Uttar Pradesh, India\\
\email{anikeitk23@iitk.ac.in}
\and
Operations and Decision Sciences,
Indian Institute of Management Ahmedabad,\\
Ahmedabad-380015, Gujarat, India\\
\email{asinha@iima.ac.in}
\and
Brij Disa Centre for Data Science and AI, 
Indian Institute of Management Ahmedabad,
Ahmedabad-380015, Gujarat, India
}

\maketitle 
\begin{abstract}
This paper proposes a Linear Programming (LP)-based local search framework for fine-tuning pretrained transformer models with explicit control against overfitting. The approach formulates transformer fine-tuning as a bilevel optimization-based regularization problem, in which model parameters and regularization hyperparameters are jointly updated. Information collected during initial warm-up iterations, including validation gradients and training Hessian information, is used to construct a local descent direction by solving an LP that minimizes a scaled directional derivative while preserving training optimality. This validation-aware descent direction enables focused local updates of both parameters and regularization hyperparameters, reducing overfitting without requiring repeated full retraining cycles. The resulting method, termed \textbf{Li}near Programming-based \textbf{F}ine-\textbf{T}uning (LiFT) for transformers, differs from conventional fine-tuning by systematically identifying task-specific updates rather than relying on heuristic or grid-based hyperparameter selection. Experiments on GPT-2 Small fine-tuned on WikiText-2 demonstrate that LiFT enables effective adaptation through selective tuning of transformer blocks and regularization parameters, yielding consistent improvements in test perplexity across multiple layer configurations and regularization settings, with particularly pronounced gains in overfitting-prone scenarios. Beyond empirical performance, LiFT establishes a principled connection between transformer fine-tuning, bilevel optimization, local search, and regularization theory.
\end{abstract}

\section{Introduction}
Large-scale transformer models have fundamentally reshaped Natural Language Processing (NLP), achieving state-of-the-art performance across a wide range of language understanding and generation tasks. Architectures such as BERT \cite{devlin2019bert} and GPT \cite{brown2020language} now serve as foundational components in many real-world systems. Adapting these models to specific tasks typically relies on fine-tuning; however, fine-tuning is most effective when large labeled datasets are available. In limited data scenarios, transformer models are prone to overfitting, often memorizing training instances rather than learning representations that generalize well. Prior studies have shown that the scale and expressiveness of modern transformers can exacerbate this issue, leading to unstable training dynamics and poor validation performance \cite{mosbach2020stability}.

To mitigate overfitting and stabilize fine-tuning, practitioners commonly employ extensive hyperparameter tuning and regularization strategies, including random search \cite{bergstra2012random}, grid search \cite{ogunsanya2023grid}, Bayesian optimization \cite{snoek2012practical}, evolutionary approaches \cite{guo2020efficient}, dropout \cite{srivastava2014dropout}, and weight decay \cite{loshchilov2017decoupled}. While effective, these techniques are computationally expensive and often require repeated cycles model training. The growing cost of fine-tuning large-scale models has significant financial and environmental implications, limiting accessibility for smaller research groups and practitioners \cite{strubell2019energy}. Moreover, in high-stakes domains such as healthcare and finance, overfitting poses serious risks, as unreliable predictions can undermine trust and lead to adverse outcomes.

Parameter-Efficient Fine-Tuning (PEFT) methods have recently emerged as a promising solution. Adapter-based approaches \cite{houlsby2019parameter} train only small additional layers while keeping the base model fixed, achieving competitive performance on benchmarks such as GLUE \cite{wang2018glue} with substantially fewer trainable parameters. LoRA \cite{hu2022lora} further reduces training complexity by introducing low-rank updates into transformer layers, enabling efficient adaptation with minimal performance loss. Prompt-based methods, including prefix-tuning \cite{li2021prefix} and soft prompt tuning \cite{lester2021power}, learn compact task-specific embeddings while updating less than 1\% of model parameters. Although these approaches significantly reduce storage and computational costs, they remain largely heuristic. Model performance can be highly sensitive to design choices such as adapter size, rank, or prompt length, often necessitating additional hyperparameter search. Importantly, these methods do not explicitly optimize for generalization, and overfitting can still occur.

\begin{figure}[ht]
\centering\includegraphics[height=5.3cm, width=8.3cm]{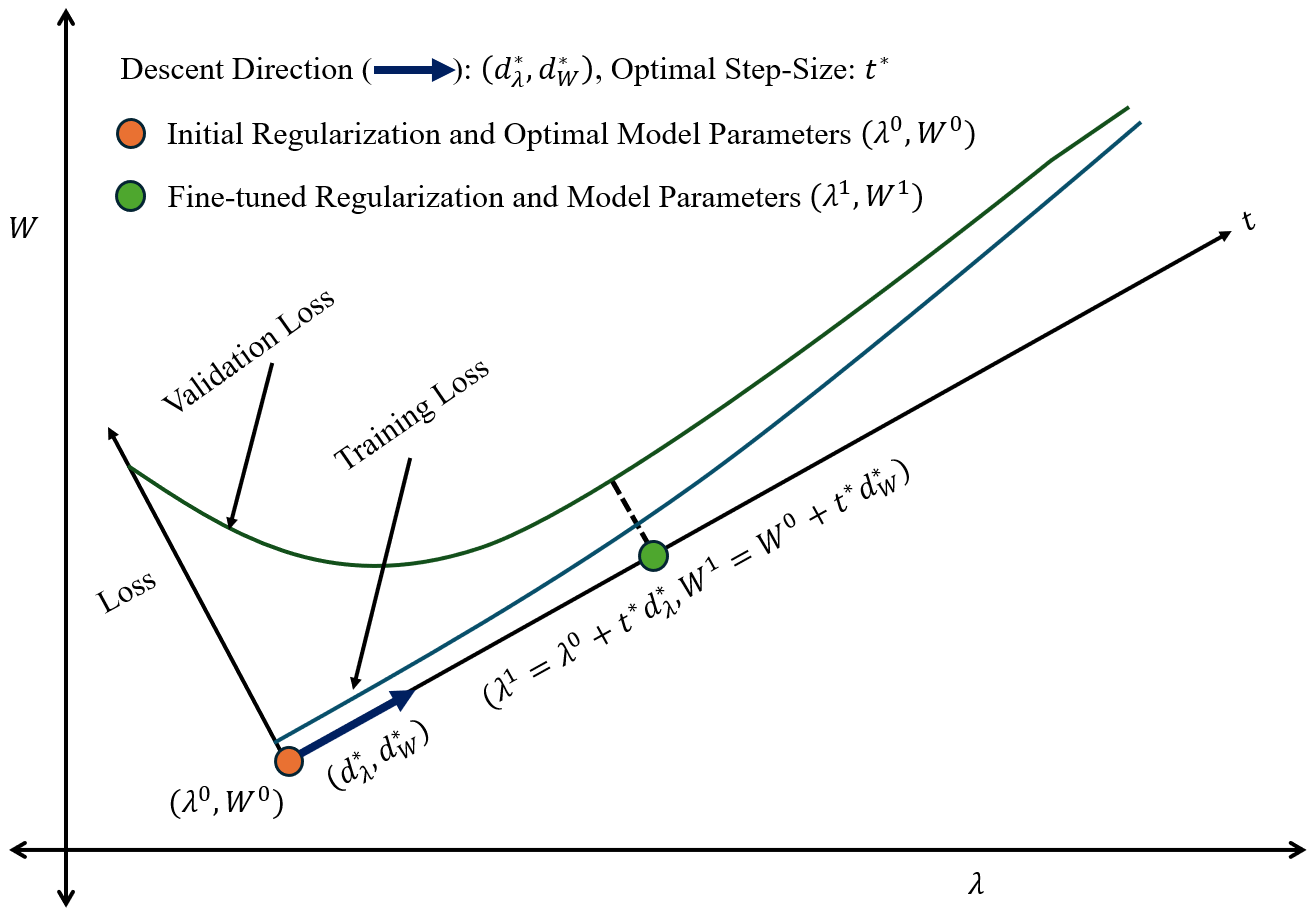}
\caption{Model fine-tuning using hyperlocal search.}
\label{fig:general_hls}
\end{figure}

To address these limitations, we propose \textbf{Li}near Programming-based \textbf{F}ine-\textbf{T}uning (LiFT) for transformers, a principled framework that explicitly incorporates validation performance into the fine-tuning process. Rather than following the training loss gradient alone, LiFT adopts a bilevel optimization-based regularization strategy~\eqref{hpo_reg} that jointly considers training optimality and validation improvement when determining the update direction. Specifically, LiFT, a first-order approach for the two-level problem, utilizes a Linear Program (LP)~\eqref{LP_l2} to compute an overfitting-aware descent direction $(d_{\lambda}^\ast, d_W^\ast)$. This direction is used to update both the regularization parameters $(\lambda)$ and the model parameters $(W)$ in the neighborhood of the current solution $(\lambda^0, W^0)$. A new solution $(\lambda^1, W^1) = (\lambda^0 + t^\ast d_\lambda^\ast, W^0 + t^\ast d_W^\ast)$ is obtained, where $t^\ast > 0$ is the optimal step along the descent direction, selected by evaluating the model $\mathcal{M}(\lambda^0 + t d_\lambda^\ast, W^0 + t d_W^\ast)$ for $t > 0$, with $t = t^\ast$ ensuring improved validation performance while preserving training optimality. This hyperlocal search process is illustrated in Figure~\ref{fig:general_hls}.

Unlike existing PEFT approaches, LiFT does not introduce additional layers or task-specific modules. Instead, it directly regularizes selected components of the transformer, including transformer blocks, embedding layers, and the language modeling head, as illustrated in Figure~\ref{fig:x_mer_hls}. Moreover, LiFT offers full flexibility over the fine-tuning configuration: any subset of parameters can be regularized while others remain frozen, allowing the tuning strategy to be tailored to available memory and computational resources. LiFT can also accommodate layer-wise fine-tuning without explicit regularization when desired. From an implementation standpoint, the method relies only on standard gradients, Hessian-vector products (when using the HVP-based LP formulation~\eqref{LP_HVP}), and an LP solver, making it lightweight and broadly applicable.  Most importantly, LiFT is not merely another heuristic, but has a theoretical motivation; it provides a principled mechanism for controlling overfitting and improving generalization, thereby reducing dependence on costly hyperparameter search.

Next, we review transformer fundamentals and related work, present the proposed fine-tuning method, and conclude with experiments, discussion, and future directions.

\section{Transformer Foundations and Past Studies on Model Fine-Tuning}
\begin{figure}[ht]
\centering
\includegraphics[height=4.75cm, width=8.5cm]{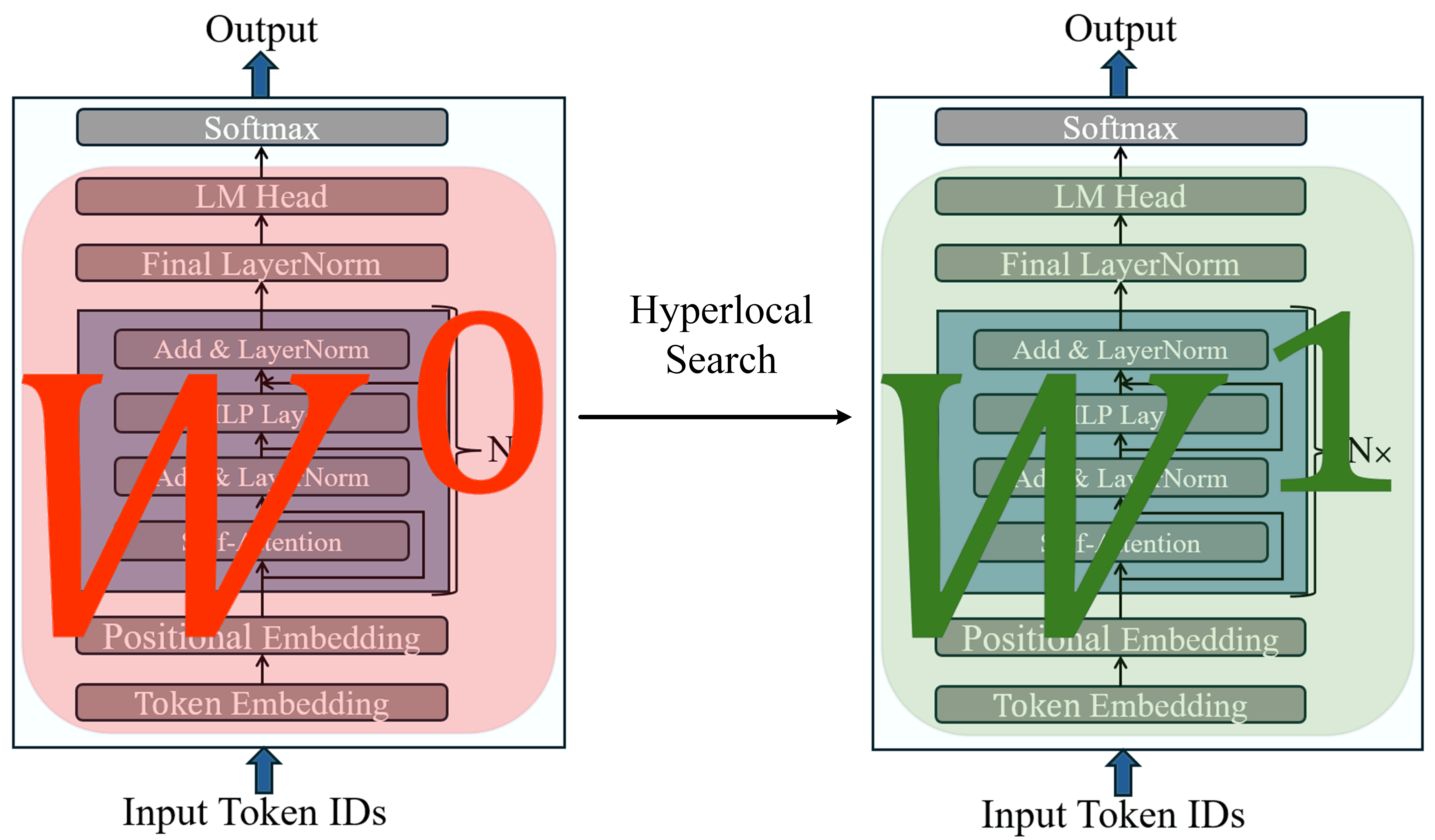}
\caption{Hyperlocal search for transformer fine-tuning.}
\label{fig:x_mer_hls}
\end{figure}

Figure~\ref{fig:x_mer_hls} provides a schematic overview of the transformer architecture. Introduced by Vaswani et al.~\cite{vaswani2017attention}, transformers revolutionized NLP by replacing recurrence and convolution with self-attention as the primary modeling mechanism. The core innovation lies in the scaled dot-product attention mechanism, which computes
attention weights between query ($Q$), key ($K$), and value ($V$) representations as
\begin{equation}
\text{Attention}(Q, K, V) = \text{softmax}\left(\frac{QK^T}{\sqrt{d_k}}\right)V,
\end{equation}
where $d_k$ denotes the dimensionality of the key vectors. This formulation enables the model to selectively focus on relevant parts of the input sequence when constructing contextualized representations.

Transformers extend this mechanism through multi-head attention, allowing the model to attend to multiple representation subspaces in parallel:
\begin{equation}
\text{MultiHead}(Q, K, V) = \text{Concat}(\text{head}_1, \ldots, \text{head}_h)W^O,
\end{equation}
where $\text{head}_i = \text{Attention}(QW^Q_i, KW^K_i, VW^V_i)$ and $W^Q_i, W^K_i, W^V_i, W^O$ are learned projection matrices. Multi-head attention enhances expressiveness by capturing diverse dependencies across positions and feature dimensions.

Each transformer block comprises two sub-layers: multi-head self-attention and a position-wise fully connected feed-forward network (FFN). The FFN applies two linear transformations with a ReLU activation, $\text{FFN}(x) = \max(0, xW_1 + b_1)W_2 + b_2$. Both sub-layers are wrapped with residual connections followed by layer normalization, expressed as $\text{LayerNorm}(x + \text{Sublayer}(x))$. This design facilitates stable optimization and efficient gradient propagation in deep architectures.
\vspace{-0.2cm}
\vspace{-0.25cm}

Following the foundational work \cite{vaswani2017attention}, transformer architectures evolved rapidly to address diverse language understanding challenges. BERT~\cite{devlin2019bert} introduced bidirectional pretraining through masked language modeling, where 15\% of input tokens are randomly masked, combined with a next sentence prediction task to capture inter-sentence relationships. This approach enabled BERT to learn rich contextual representations that could be fine-tuned for downstream tasks. The GPT series~\cite{radford2018improving,radford2019language,brown2020language} pursued an alternative autoregressive modeling approach using unidirectional attention. The original GPT demonstrated that unsupervised pretraining on large corpora followed by supervised fine-tuning could achieve strong performance across diverse tasks. GPT-2~\cite{radford2019language} dramatically scaled this approach to 1.5 billion parameters trained on 40GB of internet text, revealing emergent zero-shot and few-shot capabilities where the model could perform tasks like translation, summarization, and question answering without explicit training by conditioning on task descriptions and examples in the input context. GPT-3~\cite{brown2020language} further amplified scale to 175 billion parameters, demonstrating that sufficiently large language models can perform complex reasoning, arithmetic, and code generation through in-context learning alone, fundamentally shifting the paradigm from task-specific fine-tuning to prompt-based inference. T5~\cite{raffel2020exploring} unified tasks under a text-to-text framework, systematically studying pretraining objectives and revealing the effectiveness of denoising approaches. Then came RoBERTa~\cite{liu2019roberta} which improved BERT through refined training procedures including larger batches, dynamic masking, and removal of next sentence prediction. Subsequent innovations addressed efficiency: ELECTRA~\cite{clark2020electra} improved sample efficiency via discriminative pretraining; XLNet~\cite{yang2019generalized} captured bidirectional context through permutation language modeling; sparse attention mechanisms like Longformer~\cite{beltagy2020longformer} enabled long document processing. Despite these architectural innovations, a fundamental challenge persists: adapting large pretrained models to specific tasks through fine-tuning remains computationally expensive and prone to overfitting on small datasets.
Fine-tuning pretrained transformer models has become standard for task adaptation, but traditional approaches that update all parameters often lead to overfitting and high computational costs, particularly on smaller datasets. PEFT methods address computational costs by updating minimal parameters: Howard and Ruder~\cite{howard2018universal} introduced layer-specific learning rates in ULMFiT; adapters~\cite{houlsby2019parameter} achieve near state-of-the-art performance with 3-4\% additional parameters; prefix and prompt tuning~\cite{li2021prefix,lester2021power} optimize approximately 0.1\% of parameters; LoRA~\cite{hu2022lora} decomposes weight updates as $W_0 + BA$ with low-rank matrices, reducing trainable parameters by up to 10,000$\times$; and BitFit~\cite{zaken2022bitfit} updates only bias terms. However, PEFT methods remain heuristic, requiring manual selection of adapter dimensions or ranks without explicitly optimizing for validation performance. Regularization methods like Elastic Weight Consolidation~\cite{kirkpatrick2017overcoming}, weight decay~\cite{loshchilov2017decoupled}, and knowledge distillation~\cite{hinton2015distilling} constrain updates to prevent overfitting, but optimal strength selection relies on hyperparameter search via grid search~\cite{ogunsanya2023grid}, random search~\cite{bergstra2012random}, or Bayesian optimization~\cite{snoek2012practical}. The core limitation is decoupling hyperparameter selection from optimization: hyperparameters are fixed before training rather than jointly optimized with model parameters based on validation feedback. Bilevel optimization frames this naturally minimizing training loss at the lower-level and validation loss at the upper-level. Recent advancements in model fine-tuning reformulate the hyperparameter tuning problem as an LP using validation gradients and the training Hessian, a method referred to as Hyperlocal Search (HLS)~\cite{sinha2025linear}, and depicted in Figure~\ref{fig:general_hls}.

\section{LP-Based Local Search for Model Fine-Tuning}
In this section, we first present the HLS framework, grounded in bilevel optimization-based regularization, and derive the corresponding LP formulation. We then introduce an alternative, approximate HVP-based LP formulation that is computationally efficient and significantly faster to solve, and finally present the LiFT algorithm.

\subsection{Hyperlocal Search}
Hyperparameter optimization is commonly formulated as a bilevel optimization problem, where model training is carried out at the lower-level and generalization performance is optimized at the upper-level. By considering hyperparameters $\lambda$ and model parameters $W$, this formulation can be written as
\begin{equation}\label{hpo}
    \begin{aligned}
    \min_{\lambda, W} \quad & \mathcal{L}_v(\lambda, W) \\
    \text{s.t.} \quad & 
    W \in \argmin_{W \in \mathscr{W}} 
    \Big\{ \mathcal{L}_t(\lambda, W) \Big\}
    \end{aligned}
\end{equation}

Following prior work on regularization in machine learning, and for simplicity considering a single scalar hyperparameter together with a vector of model parameters, we reformulate the lower-level problem by introducing explicit $\ell_2$-regularization. This yields a regularized training objective in which gradient-based optimization induces an implicit regularization effect that is naturally captured by the bilevel structure. The resulting formulation is given by
\begin{equation}\label{hpo_reg}
    \begin{aligned}
    \min_{\lambda, W} \quad & \mathcal{L}_v(\lambda, W) \\
    \text{s.t.} \quad & 
    W \in \argmin_{W \in \mathscr{W}} 
    \Big\{ \mathcal{L}_t(\lambda, W) + \lambda \|W\|_{2}^{2} \Big\}
    \end{aligned}
\end{equation}

Although the above regularization problem explicitly considers a scalar $\lambda$, it can be naturally generalized to multiple hyperparameters. Since transformer architectures consist of multiple components--such as embedding layers, fully connected feed-forward networks, and multi-head attention modules--it is often desirable to associate a distinct regularization parameter with each module. Consequently, $\lambda$ can be a vector. 

Formally, let $\lambda \in \mathbb{R}^p$ denote a vector of $p$ hyperparameters and let $W \in \mathbb{R}^{n_w}$ denote the model parameters. We define the joint variable as $\theta = (\lambda, W) \in \mathbb{R}^{p+n_w}$. Here, $n_w = q \times p$, and the model parameters are partitioned as $W = [W_1, \ldots, W_p]$, where each block $W_i \in \mathbb{R}^q$ is associated with the corresponding hyperparameter $\lambda_i$. The validation loss retains the same functional form, while the regularized training loss is given by
\begin{equation}\label{reg_loss}
\mathcal{L}_t(\lambda, W) + \sum_{i=1}^{p} \lambda_i \|W_i\|_2^2 = \mathcal{L}_t(\lambda, W) + \|\sqrt{\lambda}\odot W\|_2^2,
\end{equation}
where $\odot$ denotes block-wise multiplication, i.e., $\sqrt{\lambda}\odot W := [\sqrt{\lambda_1} W_1, \ldots, \sqrt{\lambda_p} W_p]$.

To derive the LP for HLS corresponding to this bilevel regularization problem, we follow the methodology proposed in~\cite{sinha2025linear}. This leads to the following LP, which determines descent directions for both the hyperparameters and the model parameters (see Appendix~\ref{app_2a} for the complete derivation).
\begin{equation}\label{LP_l2}
    \begin{aligned}
    \min_{d_\lambda, d_W} \quad & 
    \begin{bmatrix}
        \nabla_\lambda \mathcal{L}_v(\lambda^0, W^0) \\
        \nabla_W \mathcal{L}_v(\lambda^0, W^0)
    \end{bmatrix}^{\top}
    \begin{bmatrix}
        d_\lambda \\
        d_W
    \end{bmatrix} \\
    \text{s.t.} \quad & 
    \begin{bmatrix}
        H_{21} & H_{22}
    \end{bmatrix}
    \begin{bmatrix}
        d_\lambda \\
        d_W
    \end{bmatrix} = 0, \\
    & -1 \leq d_\lambda \leq 1
    \end{aligned}
\end{equation}

Solving~\eqref{LP_l2} yields a descent direction $(d_\lambda^\ast, d_W^\ast)$. Updating the model as
$\mathcal{M}(\lambda^0 + t d_\lambda^\ast,\, W^0 + t d_W^\ast)$
guarantees a reduction, or in the worst case, no change in the validation loss while preserving lower-level training optimality. The optimal step $t^\ast$ can be determined via a line-search procedure over a suitable range of steps. Formal propositions establishing these properties of HLS are stated as Propositions~\ref{prop:lift_validation} and~\ref{prop:hls_optimality}, with their proofs provided in Appendix~\ref{app_2b}.

\begin{proposition}\label{prop:lift_validation}
(Validation Performance under HLS Update).
The HLS update obtained from \eqref{LP_l2} results to an improvement in validation performance in most cases, and in the worst case, leaves it unchanged.
\end{proposition}
\begin{proposition}\label{prop:hls_optimality}
(Preservation of Training Optimality under HLS Update).
The HLS update generated by \eqref{LP_l2} preserves optimality with respect to the training objective.
\end{proposition}

\subsection{LiFT}
The core component of the LiFT algorithm, namely HLS based on an HVP-based LP formulation, together with the LiFT algorithm itself and its time complexity, are presented as follows.\\

\noindent \tb{HVP-Based Linear Program.}
The LP formulation in the previous subsection requires explicit access to the Hessian blocks of the lower-level objective, which is not feasible for high-dimensional models. As a scalable alternative, we employ HVP to approximate the curvature constraints without forming the full Hessian explicitly.

We derive the following HVP-based LP formulation (see Appendix~\ref{app_2c} for details). Here, $H_{\text{mat}} \in \mathbb{R}^{m \times (p+n_w)}$ denotes the matrix of stacked HVPs obtained using $m$ probe vectors.
\begin{equation}\label{LP_HVP}
    \begin{aligned}
    \min_{d_\lambda, d_W} \quad & 
    \begin{bmatrix}
        \nabla_\lambda \mathcal{L}_v(\lambda^0, W^0) \\
        \nabla_W \mathcal{L}_v(\lambda^0, W^0)
    \end{bmatrix}^\top
    \begin{bmatrix}
        d_\lambda \\ d_W
    \end{bmatrix} \\
    \text{s.t.} \quad & 
    H_{\text{mat}}
    \begin{bmatrix}
        d_\lambda \\ d_W
    \end{bmatrix} = 0, \\
    & -\mathbf{1} \leq d_\lambda \leq \mathbf{1}, \quad d_W \;\text{urs},
    \end{aligned}
\end{equation}
The HVP-based LP \eqref{LP_HVP} avoids explicit Hessian construction by enforcing approximate curvature constraints along random probe directions, making the method practical for large-scale neural networks.

\noindent \tb{Hyperparameters.}
Table~\ref{tab:hyperparams_gpt} summarizes the hyperparameters employed in the fine-tuning of the transformer models.
\begin{table}[h!]
\centering
\scriptsize
\caption{Definitions of hyperparameters used in GPT-2 fine-tuning experiments.}
\label{tab:hyperparams_gpt}
\renewcommand{\arraystretch}{1.15}
\begin{tabular}{l p{0.825\columnwidth}}
\hline
\textbf{Symbol} & \textbf{Description} \\
\hline
$\mathbf{N}_{WE}$ & Number of warm-up epochs of training performed before initiating HLS \\
$\mathbb{I}_{\eta}$ & Binary indicator for including the learning rate $\eta$ in hyperparameter optimization \\
$\mathbb{I}_{\text{emb}}$ & Binary indicator for making the token embedding matrix trainable \\
$\mathbf{N}_{R}^{\text{emb}}$ & Number of regularization partitions applied to the token embedding layer; setting $\mathbf{N}_{R}^{\text{emb}}=0$ disables embedding regularization \\
$\mathbf{N}_{TB}$ & Number of transformer blocks selected for fine-tuning; if an explicit block index set is not provided, the first $\mathbf{N}_{TB}$ blocks are tuned \\
$\mathcal{B}_{\text{tune}}$ & Set of transformer block indices selected for fine-tuning, specified explicitly as a subset of all transformer block indices \\
$\mathbf{N}_{\text{FULL/MLP}}$ & Number of transformer blocks selected for fine-tuning, specifying how many are tuned fully (\texttt{FULL}) and how many only via their MLP components (\texttt{MLP}) \\
$\mathbf{N}_{R}^{\text{blk}}$ & Number of regularization partitions applied to the selected transformer block parameters; setting $\mathbf{N}_{R}^{\text{blk}}=0$ disables block-level regularization \\
$\mathbf{N}_{R}$ & Number of transformer-block regularization terms \\
$\mathbf{T}$ & Set of steps explored along the descent direction during line search \\
\hline
\end{tabular}
\end{table}

\noindent \tb{Algorithm.}
A basic version of the proposed HLS strategy for transformers (LiFT) is presented in Algorithm~\ref{liftalgo}.
\begin{algorithm}
\small
\caption{LiFT}
\label{liftalgo}
\begin{algorithmic}[1]
\setlength{\itemsep}{3pt}
\State Load the pre-trained model, initialize $\lambda^0 = [\lambda_1^0, \lambda_2^0, \ldots, \lambda_p^0]$ (initial guess), and perform $\mathbf{N}_{WE}$ epochs of warm-up training for the model parameters as follows
\[
W^0 = \argmin_{W \in \mathscr{W}} 
\left\{ \mathcal{L}_t(\lambda^0, W) + \|\sqrt{\lambda^0}\odot W\|_2^2 \right\}
\]
\State Compute validation gradients:
$\nabla_\lambda \mathcal{L}_v(\lambda^0,W^0)$ and 
$\nabla_W \mathcal{L}_v(\lambda^0,W^0)$
\State Construct HVP and formulate the HVP-based LP~\eqref{LP_HVP}
\State Solve the LP to obtain the descent direction $(d_\lambda^\ast, d_W^\ast)$
\State Perform a line search over $t \in \{t_0, t_1, \ldots, t_{|\mathbf{T}|}\}$ to generate multiple models
\[
(\lambda, W) \leftarrow (\lambda^0 + t d_\lambda^\ast, \; W^0 + t d_W^\ast)
\]
\State Evaluate performance of the above models on the validation set
\State Select $t^\ast$ minimizing validation loss and update parameters (Figures~\ref{fig:general_hls} and~\ref{fig:x_mer_hls})
\[
(\lambda^\ast, W^\ast) \leftarrow (\lambda^0 + t^\ast d_\lambda^\ast, \; W^0 + t^\ast d_W^\ast)
\]
\end{algorithmic}
\end{algorithm}
We analyze the computational (time) complexity of the proposed algorithm as follows, focusing on a single HLS iteration and its dominant components. The optimization variables $\theta = (\lambda, W) \in \mathbb{R}^{p+n_w}$ consist of continuous hyperparameters (e.g., regularization coefficients and optionally the learning rate), and selected model parameters subject to fine-tuning. Let $E_D$ be the number of training examples processed per iteration (epoch), $B$ the batch size, and $C_P$ the cost of a single forward--backward pass through the network. LiFT consists of the following computational steps. The worst case time complexity is given by
\begin{equation}
O\Bigg(
    \frac{E_D}{B} C_P\mathbf{N}_{WE} \;+\; m \frac{E_D}{B} C_P \;+\; (p+n_w)^2 \;+\; |\mathbf{T}| \frac{E_D^{(v)}}{B} C_P
\Bigg),
\label{eq:lift_complexity}
\end{equation}
which is derived in Appendix~\ref{app_2d}. In practice, the dominant computational cost arises from gradient evaluations and HVP computations, while the LP solve incurs comparatively minor overhead due to the low dimensionality of the hyperparameter space and the use of a small number of probe vectors. Importantly, all steps in LiFT admit polynomial-time complexity and avoid explicit Hessian construction, making the method scalable to large transformer models under selective fine-tuning regimes. Although Algorithm~\ref{liftalgo} describes a single iteration of HLS, it can be applied multiple times, yielding the sequence $(\lambda^0, W^0) \rightarrow (\lambda^1, W^1) \rightarrow (\lambda^2, W^2) \rightarrow \cdots$ with progressively improved generalization performance. However, this comes at the cost of increased computational complexity, since HLS must be executed multiple times.

\section{Experimental Results and Discussion}
In this section, we present the results of LiFT on the WikiText-2 dataset for fine-tuning the GPT-2 Small base model and discuss the main experimental findings.
\subsection{Dataset and Baseline Model}
All experiments are conducted on the WikiText-2 dataset, a widely used corpus for evaluating language modeling and fine-tuning strategies on transformer-based architectures. WikiText-2 consists of curated Wikipedia articles with minimal preprocessing, preserving long-range dependencies and natural linguistic structure. Following standard practice, we partition the dataset into training, validation, and test splits, containing $4{,}718$, $487$, and $558$ fixed-length sequences, respectively.

As the baseline model, we employ the GPT-2 Small architecture, comprising a total of $12$ transformer blocks and $124.4$ million parameters. The model is initialized from pretrained weights and retains its original token and positional embedding structure. No architectural modifications are introduced at initialization; instead, fine-tuning is restricted to carefully selected subsets of pretrained parameters. This design choice allows us to directly attribute performance changes to LiFT’s parameter selection and bilevel updates, rather than to architectural expansions or reparameterization effects.

\subsection{Experimental Setup}
We evaluate the proposed LiFT strategy in a constrained fine-tuning regime, where only a small subset of transformer blocks is adapted while all remaining network parameters are kept frozen. All experiments are conducted on a High-Performance Computing (HPC) cluster using a single node equipped with two GPUs and sufficient system memory to support GPT-2 fine-tuning. A warm-up phase is performed using the Adam optimizer with a fixed learning rate ($\mathbb{I}_{\eta}$=0, $\eta=10^{-4}$), and gradient clipping is applied to ensure numerical stability. A base regularization strength of $10^{-4}$ is used throughout the experiments. In all configurations considered, embedding parameters are neither tuned nor regularized ($\mathbb{I}_{\text{emb}}=\mathbf{N}_{R}^{\text{emb}}=0$). Instead, block-level regularization is enabled via $\ell_2$ penalties applied to disjoint partitions of the selected transformer block parameters. Training is carried out with a batch size of $4$. All input sequences are tokenized using the GPT-2 tokenizer and truncated or padded to a maximum length of $512$ tokens, enabling efficient batched training while maintaining sufficient contextual coverage. For computing validation gradients, we use at most $10$ batches of data. HVP constraints are approximated using $m=16$ probe vectors with a finite-difference step-size of $\varepsilon = 10^{-3}$. The resulting LPs which are convex by nature, are solved using the open-source \texttt{CVXPY} Python package. Finally, a line search is performed over $|\mathbf{T}|$ candidate steps ($<=1.5$) in each experiment.

\subsection{Fine-tuning Results and Discussion}
\begin{figure}[ht]
\centering
\includegraphics[height=8.5cm, width=12cm]{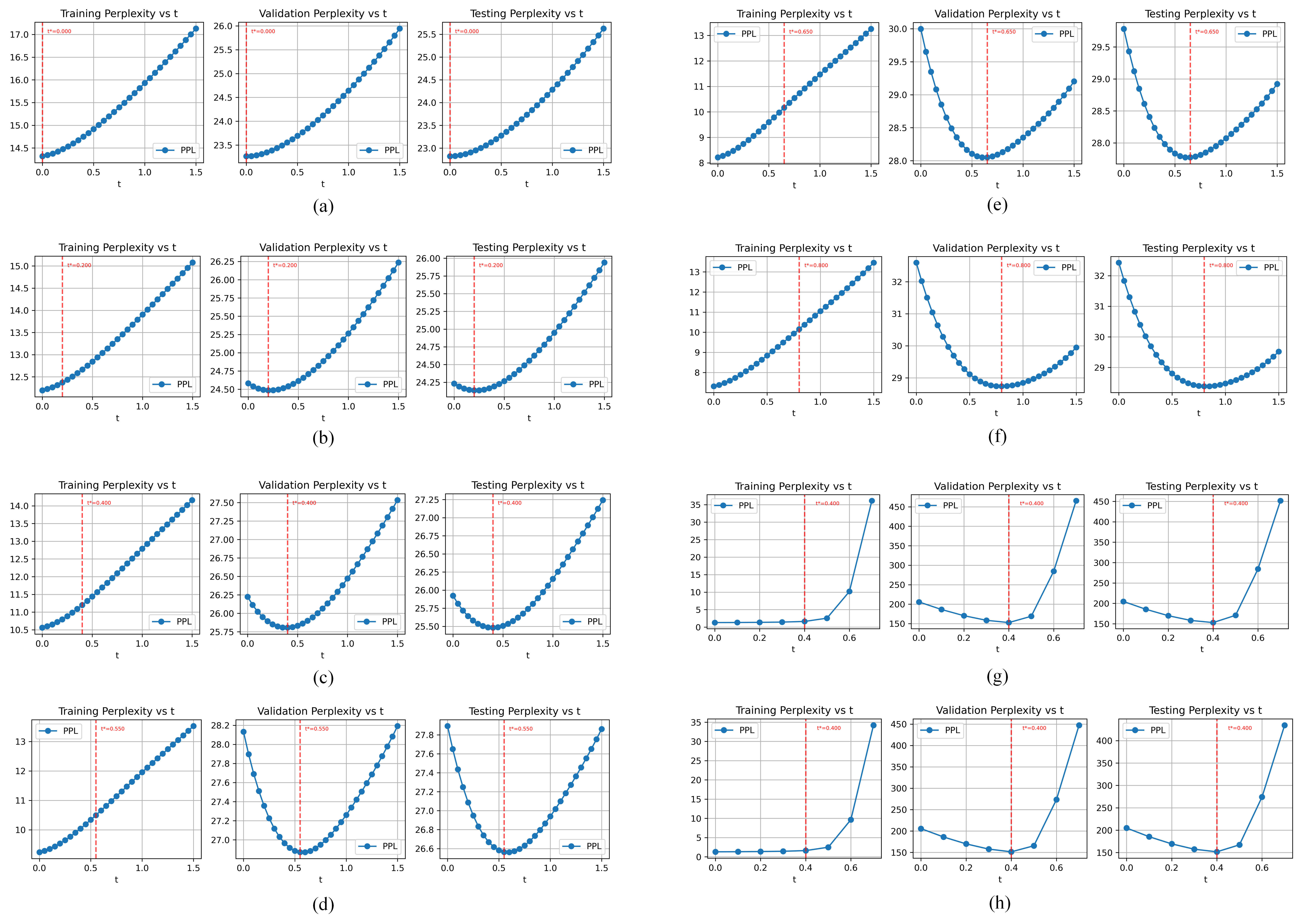}
\caption{Hyperlocal search for optimal step identification in GPT-2 fine-tuning with training, validation, and test perplexity curves.}
\label{fig:hls}
\end{figure}

For different hyperparameter configurations, Table~\ref{tab:lift} reports the corresponding training, validation, and test perplexity (together with the associated loss values), and Figure~\ref{fig:hls} depicts the validation loss evaluated over a range of steps for the considered hyperparameter configurations, with the optimal step selected as the one minimizing the validation perplexity. The effectiveness of our algorithm in controlling overfitting and improving model generalization is clearly demonstrated by the experimental results. In particular, we observe an increase in training perplexity alongside a substantial reduction in both validation and test perplexities.
\begin{table}[h!]
\centering
\scriptsize
\caption{Training, validation, and test perplexity (PPL)/loss for WikiText-2 dataset with and without HLS.\\}
\label{tab:lift}
\resizebox{\columnwidth}{!}{%
\begin{tabular}{llllllccc}
\hline
\multicolumn{4}{c}{\textbf{HP}} & \textbf{\shortstack{Sample \\ Size}} & \textbf{\shortstack{PPL/ \\ Loss}} & \textbf{\shortstack{Without \\ HLS}} & \textbf{\shortstack{With \\ HLS}} & \textbf{\shortstack{Test PPL \\ $\downarrow$\;(\%)}} \\
\cline{1-4}
\(\mathbf{N}_{WE}\) & \(\mathbf{N}_{TB}\) & \(\mathbf{N}_{R}\) & \(\mathbf{N}_{\text{FULL/MLP}}\)
 & & & & \\
\hline
 & & & &  4718 & Training   & 14.32/2.66  & 14.32/2.66  \\
 10 & 3 & 2 & 1/2 &  487  & Validation & 23.26/3.15  & 23.26/3.15 \\
 & & & &  558  & Testing    & 22.82/3.13  & 22.82/3.13 & 0.00 \\
\hline
 & & & &  4718 & Training   & 12.20/2.50  & 12.37/2.52 \\
 15 & 3 & 2 & 1/2 &  487  & Validation & 24.58/3.20  & 24.48/3.20 \\
 & & & &  558  & Testing    & 24.23/3.19  & 24.14/3.18 & 0.37 \\
\hline
 & & & &  4718 & Training   & 10.56/2.36  & 11.20/2.42 \\
 20 & 3 & 2 & 1/2 &  487  & Validation & 26.22/3.27  & 25.81/3.25  \\
 & & & &  558  & Testing    & 25.92/3.26   & 25.48/3.24 & 1.70 \\
\hline
 & & & &  4718 & Training   & 9.24/2.22  & 10.50/2.35 \\
 25 & 3 & 2 & 1/2 &  487  & Validation & 28.13/3.34  & 26.87/3.29 \\
 & & & &  558  & Testing    & 27.89/3.33  & 26.56/3.28 & 4.77 \\
\hline
 & & & &  4718 & Training   & 8.21/2.11  & 10.17/2.32 \\
 30 & 3 & 2 & 1/2 &  487  & Validation & 30.00/3.40  & 28.05/3.33 \\
 & & & &  558  & Testing    & 29.78/3.39  & 27.77/3.32 & 6.75 \\
\hline
 & & & &  4718 & Training   & 7.31/1.99   & 10.16/2.32 \\
 35 & 3 & 2 & 1/2 &  487  & Validation & 32.60/3.48  & 28.73/3.36  \\
 & & & &  558  & Testing    & 32.43/3.48   & 28.38/3.35 & 12.49\\
\hline
 & & & &  4718 & Training   & 1.30/0.26  & 1.62/0.48 \\
 25 & 4 & 4 & 2/2 &  487  & Validation & 205.62/5.33   & 152.73/5.03\\
 & & & &  558  & Testing    & 204.86/5.32   & 153.03/5.03 & 25.30 \\
\hline
 & & & &  4718 & Training   & 1.30/0.26  & 1.61/0.48 \\
 25 & 5 & 5 & 2/3 &  487  & Validation & 205.62/5.33 & 151.38/5.02 \\
 & & & &  558  & Testing    & 204.86/5.32 & 151.71/5.02 & 25.94 \\
\hline
\end{tabular}%
}
\end{table}

To illustrate one of the reported experiments, consider the fine-tuning is applied to $3$ ($\mathbf{N}_{TB}$) non-consecutive transformer block layers $\{1,7,11\}$ ($\mathcal{B}_{\text{tune}}$). For these blocks, heterogeneous tuning modes are considered: the full parameter set is optimized for block $7$, while only the MLP components are tuned for blocks $1$ and $11$ ($\mathbf{N}_{\text{FULL/MLP}} = 1/2$), with their regularization configurations given by $\mathbf{N}_{R}^{1} = 1$, $\mathbf{N}_{R}^{7} = 0$, $\mathbf{N}_{R}^{11} = 1$, and $\mathbf{N}_{R} = 2$. This mixed structural selection reflects practical scenarios in which different layers exhibit varying sensitivity to task-specific adaptation. Prior to invoking HLS, the selected parameters undergo a warm-up phase of $35$ ($\mathbf{N}_{WE}$) epochs of standard gradient-based training. This stage stabilizes the optimization landscape and provides a meaningful initialization for the subsequent bilevel procedure. Following warm-up, a single iteration of LP-based HLS is performed. At this stage, a descent direction is computed by solving an LP constructed using randomized HVP-based constraints. A line search is then carried out over a predefined set of steps ($|\mathbf{T}|=31$), and the step yielding the lowest validation loss is selected. In the reported run, the optimal step is $0.8$ ($t^\star$), determined solely based on validation performance. The updated parameters are subsequently evaluated on the training, validation, and test datasets using standard language modeling metrics, including loss and perplexity. In this experiment, LiFT improved the test perplexity by approximately $12.5\%$ compared to the baseline GPT-2 model. Across the $8$ experimental runs, the end-to-end runtime per experiment---including warm-up, HLS, and evaluation---ranges from approximately $1.59$ to $3.45$ hours, with an average runtime of about $2.54$ hours. Intermediate results, summary statistics, and diagnostic plots are logged automatically to facilitate post-hoc analysis and reproducibility.

Each experiment begins with a warm-up training phase, which naturally encourages overfitting to the training data. This phase is intentional and essential, as it produces a WikiText-2 overfitted GPT-2 model on which LiFT can be meaningfully applied. Following this phase, LiFT successfully controls overfitting (reflected in higher training perplexity) while simultaneously enhances generalization performance (reflected in lower validation and test perplexities). Importantly, when a model does not exhibit overfitting and already generalizes well, further fine-tuning is unnecessary. In such cases, the model’s performance is already satisfactory, and LiFT appropriately refrains from modifying the parameters (first entry in Table~\ref{tab:lift}). This behavior is further confirmed by additional experiments reported in Appendix~\ref{app_3}. Overall, the tabulated results and supplementary experiments consistently show that LiFT prescribes updates only in directions that improve generalization while controlling overfitting. When the GPT-2 model is already optimal and free from overfitting, the algorithm correctly recommends no parameter updates, thereby avoiding unnecessary or misguiding modifications to the model. We further observe that the results from the last two experiments indicate that substantially higher improvements in test performance-up to approximately \(25.9\%\)—can be achieved for highly overfitted GPT-2 models. However, despite these gains, such models may still be of limited practical utility, as they continue to exhibit overfitting and therefore require additional iterations of HLS to achieve improved generalization. Figure~\ref{fig:ppl_red} in Appendix~\ref{app_3} shows that LiFT is increasingly recommended as overfitting intensifies with additional warm-up epochs.

In many fine-tuning and hyperparameter optimization methods, such as random search, grid search, and evolutionary computation–based hyperparameter search, updates to hyperparameters (including architectural choices, optimization settings, and regularization parameters) are typically decoupled from updates to the model parameters. In contrast, our approach explicitly enforces synergy between these two levels: each hyperparameter update is accompanied by a corresponding optimal update of the model parameters. While a single iteration of HLS may not yield globally optimal hyperparameters--due to its dependence on the initial hyperparameter guess and its inherently local nature--repeated HLS iterations allow the method to progressively approach optimal hyperparameter configurations.

Modern deep neural networks are composed of layers that serve fundamentally different representational and computational roles, and therefore exhibit distinct optimization and generalization behaviors. Early layers tend to learn generic, low-level features such as edges or frequency components, whereas deeper layers capture higher-level, task-specific abstractions that are more sensitive to noise. As a result, the distribution of weights, gradient magnitudes, and susceptibility to overfitting can vary substantially across layers. Applying a uniform learning rate or regularization strength across all layers therefore ignores these intrinsic differences and is often suboptimal in practice. For example, early layers-shared across many downstream computations-benefit from weaker regularization to preserve representational flexibility, while later layers, which directly influence the loss, often require stronger regularization to control overfitting. This motivates the need for layer-specific hyperparameter updates, as opposed to uniform updates applied across all layers. Empirical studies, including~\cite{snoek2012practical}, indicate that treating layer-wise hyperparameters (such as distinct weight cost per layer) as tunable can materially affect performance, although the specific effect of using separate regularization parameters across layers has not been systematically analyzed. The present work also aims to address this gap by enabling principled, layer-wise hyperparameter optimization within a bilevel optimization-based regularization framework.

\section{Conclusions and Future Scope}

This paper introduced LiFT, an optimization-driven alternative to conventional fine-tuning strategies for pretrained transformer models. In contrast to heuristic update rules and costly hyperparameter search, LiFT adopts a local search perspective grounded in bilevel optimization to guide fine-tuning in a principled and computationally efficient manner. By explicitly incorporating validation behavior into the update direction, the proposed framework enables controlled adaptation of large models and mitigates overfitting without requiring repeated full retraining. Experimental results on GPT-2 Small fine-tuned on WikiText-2 demonstrate that optimization-guided updates can consistently improve generalization across different layer selections and regularization configurations.

Empirically, LiFT achieves an improvement of approximately $12.5\%$ in test perplexity over the baseline GPT-2 model (see the $6^{\text{th}}$ entry in Table~\ref{tab:lift}), with gains becoming more pronounced as overfitting increases. Beyond these results, LiFT provides a broader conceptual contribution by connecting transformer fine-tuning with bilevel optimization, local search, and regularization theory. This perspective reframes fine-tuning as a structured optimization problem rather than a collection of ad hoc design choices. From a practical standpoint, LiFT offers a lightweight and flexible mechanism for adapting large models under limited computational budgets, while maintaining reliability in deployment settings where generalization is critical. It should be noted that the computational complexity associated with Hessian computation has been mitigated through an HVP-based LP formulation that uses only a few affine constraints.

Several avenues for future work arise from this study. Fine-tuning methods such as LoRA and PEFT do not involve hyperparameter updates; LiFT can be readily combined with them to improve robustness and generalization. Future work includes extending LiFT to larger transformer architectures, applying it to domain-specific training while mitigating overfitting, and exploring adaptive trust-region or constraint-scaling strategies within the LP formulation. More broadly, LiFT suggests a promising direction in which fine-tuning is treated as a principled optimization problem, enabling more generalization-aware and theoretically grounded training of large neural networks.

\bibliographystyle{splncs04}
\bibliography{references}
\clearpage
\appendix

\section{LP-Based Local Search}
\subsection{Hyperlocal Search}\label{app_2a}
Here we assume that for any given $\lambda$, the lower-level solution exists, the validation loss is continuously differentiable, and the training loss with $\ell_2$-regularization is twice differentiable. Expanding the validation loss around $(\lambda^0, W^0)$ using a first-order Taylor's approximation yields
\begin{equation*}
    \begin{aligned}
    \mathcal{L}_v(\lambda^0 + t d_\lambda, W^0 + t d_W) =  \ \mathcal{L}_v(\lambda^0, W^0) + t \langle \nabla_\lambda \mathcal{L}_v(\lambda^0, W^0), d_\lambda \rangle + \\
    t \langle \nabla_W \mathcal{L}_v(\lambda^0, W^0), d_W \rangle,
    \end{aligned}
\end{equation*}
where $t>0$ is the step and $[d_\lambda, d_W]^T$ is the update direction. A decrease in validation loss is guaranteed if
\[
\left\langle 
\begin{bmatrix}
    \nabla_\lambda \mathcal{L}_v(\lambda^0, W^0) \\
    \nabla_W \mathcal{L}_v(\lambda^0, W^0)
\end{bmatrix},
\begin{bmatrix}
    d_\lambda \\
    d_W
\end{bmatrix}
\right\rangle < 0
\]
The update must also preserve the optimality of model parameters under perturbations in $\lambda$. This leads to the condition
\begin{equation}\label{lowerleveldir_l2}
    d_W \in \argmin_{d_W} 
    \left\{
    \begin{aligned}
        & \mathcal{L}_t(\lambda^0 + t d_\lambda, W^0 + t d_W) \\
        & + \|\sqrt{\lambda^0 + t d_\lambda}\odot (W^0 + t d_W)\|_2^2
    \end{aligned}
    \right\}
\end{equation}
Approximating the lower-level objective quadratically at $(\lambda^0, W^0)$ gives
\begin{equation}\label{Taylor_l2}
    \begin{aligned}
    \mathcal{L}_t(\lambda^0 + t d_\lambda, W^0 + t d_W) 
    + \|\sqrt{\lambda^0 + t d_\lambda}\odot (W^0 + t d_W)\|_2^2
    =  \ \mathcal{L}_t(\lambda^0, W^0) + \|\sqrt{\lambda^0 }\odot W^0\|_2^2 \\
    + t (\langle \nabla_\lambda (\mathcal{L}_t+\|\sqrt{\lambda }\odot W\|_2^2)(\lambda^0,W^0), d_\lambda \rangle + \langle \nabla_W (\mathcal{L}_t+\|\sqrt{\lambda }\odot W\|_2^2)(\lambda^0,W^0), d_W \rangle)\\
    + \tfrac{1}{2}t^2 
    \begin{bmatrix}
        d_\lambda \\
        d_W
    \end{bmatrix}^\top 
    \nabla^2_{(\lambda,W)} (\mathcal{L}_t+\|\sqrt{\lambda }\odot W\|_2^2)(\lambda^0,W^0) 
    \begin{bmatrix}
        d_\lambda \\
        d_W
    \end{bmatrix}
    \end{aligned}
\end{equation}
At optimality, the linear terms in $d_W$ vanish, reducing \eqref{lowerleveldir_l2} to
\begin{equation}\label{lowerleveldir_quad}
    d_W \in \argmin_{d_W}
    \begin{bmatrix}
        d_\lambda \\
        d_W
    \end{bmatrix}^\top
    \nabla^2_{(\lambda,W)} (\mathcal{L}_t+\|\sqrt{\lambda }\odot W\|_2^2)(\lambda^0,W^0)
    \begin{bmatrix}
        d_\lambda \\
        d_W
    \end{bmatrix}
\end{equation}
Considering the Hessian blocks as
\begin{equation}\label{hessian_matrix}
    \nabla^2_{(\lambda,W)} (\mathcal{L}_t+\|\sqrt{\lambda }\odot W\|_2^2)(\lambda^0,W^0)
=
\begin{bmatrix}
    H_{11} & H_{12} \\
    H_{21} & H_{22}
\end{bmatrix},
\end{equation}
the first-order optimality condition of lower-level problem becomes
\begin{equation}\label{firstorder_l2}
    \begin{bmatrix}
        H_{21} & H_{22}
    \end{bmatrix}
    \begin{bmatrix}
        d_\lambda \\
        d_W
    \end{bmatrix} = 0
\end{equation}
To obtain the steepest descent direction with unit-norm scaling, we arrive at the following Second Order Conic Program (SOCP)
\begin{equation}\label{SOCP_l2}
    \begin{aligned}
    \min_{d_\lambda, d_W} \quad & 
    \begin{bmatrix}
        \nabla_\lambda \mathcal{L}_v(\lambda^0, W^0) \\
        \nabla_W \mathcal{L}_v(\lambda^0, W^0)
    \end{bmatrix}^\top
    \begin{bmatrix}
        d_\lambda \\
        d_W
    \end{bmatrix} \\
    \text{s.t.} \quad & 
    \begin{bmatrix}
        H_{21} & H_{22}
    \end{bmatrix}
    \begin{bmatrix}
        d_\lambda \\
        d_W
    \end{bmatrix} = 0, \\
    & \|d_\lambda\|_2 \leq 1
    \end{aligned}
\end{equation}
Relaxing this SOCP to an LP yields
\begin{equation}
    \begin{aligned}
    \min_{d_\lambda, d_W} \quad & 
    \begin{bmatrix}
        \nabla_\lambda \mathcal{L}_v(\lambda^0, W^0) \\
        \nabla_W \mathcal{L}_v(\lambda^0, W^0)
    \end{bmatrix}^\top
    \begin{bmatrix}
        d_\lambda \\
        d_W
    \end{bmatrix} \\
    \text{s.t.} \quad & 
    \begin{bmatrix}
        H_{21} & H_{22}
    \end{bmatrix}
    \begin{bmatrix}
        d_\lambda \\
        d_W
    \end{bmatrix} = 0, \\
    & -1 \leq d_\lambda \leq 1
    \end{aligned}
\end{equation}

\subsection{Validation Improvement (Proposition~1) and Lower-Level Optimality Preservation (Proposition~2) under HLS Update}\label{app_2b}

\paragraph{Proof of Proposition~1.} The point $(d_\lambda, d_W) = (0,0)$ is feasible for~\eqref{LP_l2} and attains an objective value of zero, since it satisfies both the affine and box constraints. As~\eqref{LP_l2} is a minimization problem, any feasible solution provides a valid upper bound on the optimal value. Consequently, the optimal solution $(d_\lambda^*, d_W^*)$ always has an objective value no greater than zero. Hence, the HLS update in most cases, leads to an improvement in validation performance, and under the worst case, it leads to no change.
\hfill $\blacksquare$

\paragraph{Proof of Proposition~2.} Let the model parameters \(W^0\) be optimal for the lower-level problem in~\eqref{hpo_reg} corresponding to given regularization parameter \(\lambda^0\in \mathbb{R}^p\). The affine (linear) constraints of the LP~\eqref{LP_l2} can be expressed as
\begin{equation}\label{affine_l2}
    H_{21} d_\lambda + H_{22} d_W = 0
\end{equation}
Using the quadratic Taylor expansion~\eqref{Taylor_l2} together with the Hessian block structure in~\eqref{hessian_matrix}, we obtain
\begin{equation}\label{Taylor_l2_1}
    \begin{aligned}
\mathcal{L}_t(\lambda^0 + t d_\lambda, W^0 + t d_W) 
    + \|\sqrt{\lambda^0 + t d_\lambda}\odot (W^0 + t d_W)\|_2^2
    =  \ \mathcal{L}_t(\lambda^0, W^0) + \|\sqrt{\lambda^0 }\odot W^0\|_2^2 \\
    + t (\langle \nabla_\lambda (\mathcal{L}_t+\|\sqrt{\lambda }\odot W\|_2^2)(\lambda^0,W^0), d_\lambda \rangle + \langle \nabla_W (\mathcal{L}_t+\|\sqrt{\lambda }\odot W\|_2^2)(\lambda^0,W^0), d_W \rangle)\\
    + \tfrac{1}{2}t^2 
    \begin{bmatrix}
        d_\lambda \\
        d_W
    \end{bmatrix}^\top 
\begin{bmatrix}
H_{11} & H_{12} \\
H_{21} & H_{22}
\end{bmatrix}
    \begin{bmatrix}
        d_\lambda \\
        d_W
    \end{bmatrix}
    \end{aligned}
\end{equation}
Invoking the affine constraint~\eqref{affine_l2}, the quadratic term simplifies, yielding
\begin{equation}\label{Taylor_l2_2}
    \begin{aligned}
    \mathcal{L}_t(\lambda^0 + t d_\lambda, W^0 + t d_W) 
    + \|\sqrt{\lambda^0 + t d_\lambda}\odot (W^0 + t d_W)\|_2^2
    =  \ \mathcal{L}_t(\lambda^0, W^0) + \|\sqrt{\lambda^0 }\odot W^0\|_2^2 \\
    + t (\langle \nabla_\lambda (\mathcal{L}_t+\|\sqrt{\lambda }\odot W\|_2^2)(\lambda^0,W^0), d_\lambda \rangle + \langle \nabla_W (\mathcal{L}_t+\|\sqrt{\lambda }\odot W\|_2^2)(\lambda^0,W^0), d_W \rangle)\\
    + \tfrac{1}{2}t^2 \left\langle 
d_\lambda, \big(H_{11} - H_{12}H_{22}^{-1}H_{21}\big)d_\lambda
\right\rangle
\end{aligned}
\end{equation}
From~\eqref{affine_l2}, the update direction \(d_\lambda\) can be expressed in terms of \(d_W\) by assuming the pseudo-inverse of \(H_{21}\), denoted by \(H_{21}^\dagger\),
\begin{equation}\label{d_l_d_w}
    d_\lambda = -H_{21}^\dagger H_{22} d_W
\end{equation}
Substituting~\eqref{d_l_d_w} into~\eqref{Taylor_l2_2} yields the second-order gradient, or equivalently the new Hessian matrix, with respect to the updated model parameters \((W^0 + t d_W)\),
\begin{equation*}\label{newH}
   H'' = (H_{21}^\dagger H_{22})^T(H_{11} - H_{12}H_{22}^{-1}H_{12}^T)H_{21}^\dagger H_{22}
\end{equation*}
Here, \(S = H_{11} - H_{12}H_{22}^{-1}H_{12}^T\) denotes the Schur complement of \(H_{22}\) in the original Hessian matrix~\eqref{hessian_matrix}. If the original Hessian matrix is Positive Semidefinite (PSD), then its Schur complement \(S\) is also PSD. Consequently, \(H''\) is PSD if \(z^T H'' z \geq 0\) for all \(z \in \mathbb{R}^{n_w}\). To verify this, consider an arbitrary vector \(y \in \mathbb{R}^{n_w}\),
\[
y^T H'' y = y^T \big((H_{21}^\dagger H_{22})^T S (H_{21}^\dagger H_{22})\big) y
\]
Let \(v = H_{21}^\dagger H_{22} y\), we obtain $y^T H'' y = v^T S v$. Since \(S\) is PSD, it follows that \(v^T S v \geq 0\) for all \(v \in \mathbb{R}^p\). Therefore, $y^T H'' y \geq 0 \quad \forall y \in \mathbb{R}^{n_w}$, which confirms that \(H''\) is positive semidefinite.

In conclusion, the positive semidefiniteness of the new Hessian matrix \(H''\) implies that the HLS update remains confined to the lower-level optimality region. As a result, for sufficiently small \(t>0\), the perturbed parameters \(W^0 + t d_W\) remain optimal solutions of the lower-level problem corresponding to hyperparameter vector \(\lambda^0 + t d_\lambda\). Hence, the proposed LP update preserves lower-level optimality under the $\ell_2$-regularized bilevel formulation. \hfill $\blacksquare$

\subsection{HVP-Based Linear Program}\label{app_2c}
The update direction is $d = (d_\lambda, d_W) \in \mathbb{R}^{p+n_w}$. In the devised LP~\eqref{LP_l2}, the descent direction is constrained by
\[
\nabla^2_{(\lambda,W)} \!\left( \mathcal{L}_t(\lambda,W) + \|\sqrt{\lambda}\odot W\|_2^2 \right)\!(\lambda^0,W^0)
\begin{bmatrix}
d_\lambda \\ d_W
\end{bmatrix}
= 0,
\]
which requires the computation of the full Hessian matrix. To approximate these constraints efficiently, we introduce $m$ probe vectors $\{v^{(1)}, \ldots, v^{(m)}\} \subset \mathbb{R}^{p+n_w}$ and compute the corresponding HVPs
\[
h^{(i)} \;=\; 
\nabla^2_{(\lambda,W)} \!\left( \mathcal{L}_t(\lambda,W) + \|\sqrt{\lambda}\odot W\|_2^2 \right)\!(\lambda^0,W^0)\, v^{(i)}, 
\qquad i=1,\ldots,m
\]
For each probe we impose the orthogonality condition
\[
(h^{(i)})^\top 
\begin{bmatrix}
d_\lambda \\ d_W
\end{bmatrix} = 0
\]
Stacking them yields
\[
H_{\text{mat}}
\begin{bmatrix}
d_\lambda \\ d_W
\end{bmatrix} = 0, 
\quad H_{\text{mat}} \in \mathbb{R}^{m \times (p+n_w)},
\]
where the $i$-th row is $(h^{(i)})^\top$. The resulting HVP-based LP formulation is
\begin{equation}
    \begin{aligned}
    \min_{d_\lambda, d_W} \quad & 
    \begin{bmatrix}
        \nabla_\lambda \mathcal{L}_v(\lambda^0, W^0) \\
        \nabla_W \mathcal{L}_v(\lambda^0, W^0)
    \end{bmatrix}^\top
    \begin{bmatrix}
        d_\lambda \\ d_W
    \end{bmatrix} \\
    \text{s.t.} \quad & 
    H_{\text{mat}}
    \begin{bmatrix}
        d_\lambda \\ d_W
    \end{bmatrix} = 0, \\
    & -\mathbf{1} \leq d_\lambda \leq \mathbf{1}, \quad d_W \;\text{urs},
    \end{aligned}
\end{equation}
where inequalities are component-wise on the $p$-dimensional vector $d_\lambda$.\\

\noindent \textbf{HVP computation.} Each HVP $h^{(i)}$ is computed without explicitly forming the Hessian. For a probe $v \in \mathbb{R}^{p+n_w}$ and a small $\varepsilon > 0$, we approximate
\[
h \;\approx\; \frac{\nabla_\theta \mathcal{L}_t(\theta^0+\varepsilon v) - \nabla_\theta \mathcal{L}_t(\theta^0)}{\varepsilon},
\]
with $\theta^0 = (\lambda^0, W^0)$. The computation steps are given as follows
\begin{enumerate}
    \item Save current parameters $\theta^0$
    \item Perturb: $\theta^+ = \theta^0 + \varepsilon v$
    \item Compute $g^+ = \nabla_\theta \mathcal{L}_t(\theta^+)$
    \item Restore $\theta^0$ and compute $g = \nabla_\theta \mathcal{L}_t(\theta^0)$
    \item Form $h \approx (g^+ - g)/\varepsilon$
\end{enumerate}
Each probe requires only two gradient evaluations, making the method efficient. In practice, $m \ll p+n_w$, with typical values $m \in \{4,8,16\}$, and probe vectors are normalized random directions.

\subsection{Time Complexity}\label{app_2d}
\begin{enumerate}
    \item \textbf{Warm-up and gradient computations.}  
    Prior to HLS, a warm-up phase of gradient-based training is performed on the selected parameters. For a fixed number of warm-up epochs ($\mathbf{N}_{WE}$), the cost is
    \[
    O\!\left(\frac{E_D}{B} C_P\mathbf{N}_{WE} \right)
    \]
    During HLS, computing validation gradients with respect to $(\lambda,W)$ also incurs linear cost in the number of parameters, i.e., $O(p+n_w)$ per batch, which is dominated by forward--backward passes.

    \item \textbf{HVP approximation.}  
    Each HVP is computed using a finite-difference approximation requiring two gradient evaluations of the lower-level objective. Hence, the cost of a single HVP is
    \[
    O\!\left(\frac{E_D}{B} C_P \right)
    \]
    Using $m$ probe vectors, the total cost of constructing the HVP-based constraint matrix is
    \[
    O\!\left(m \frac{E_D}{B} C_P \right)
    \]
    \item \textbf{LP formulation and solution.}  
    The resulting LP involves $p+n_w$ decision variables and $m$ linear equality constraints, along with box constraints on the $p$ hyperparameter directions. In the worst case, generic LP solvers exhibit polynomial-time complexity. In practice, the effective computational cost is well approximated by quadratic complexity. Although interior-point methods admit worst-case cubic bounds, the LPs arising in our setting are low-dimensional in the hyperparameter space and include only a small number of HVP-induced equality constraints relative to the total number of variables. Moreover, the constraint matrix is dense but highly structured, consisting primarily of linear orthogonality constraints, which further facilitates efficient solution. As a result, the empirical cost of solving the LP is better characterized as
    \[
    O((p+n_w)^2),
    \]
    which is typically negligible compared to gradient and HVP computations for large neural networks.
    \item \textbf{Line search and evaluation.}  
    A set of candidate steps $\{t_1,\ldots,t_{|\mathbf{T}|}\}$ is evaluated by forward passes on the validation set. This incurs a cost of
    \[
    O\!\left(|\mathbf{T}| \frac{E_D^{(v)}}{B} C_P \right),
    \]
    where $E_D^{(v)}$ denotes the number of validation examples. Since $|\mathbf{T}|$ is small and fixed in practice, this term does not dominate the overall complexity.
\end{enumerate}
Combining the above components, the total time complexity of LiFT is
\begin{equation*}
O\Bigg(
    \frac{E_D}{B} C_P\mathbf{N}_{WE} \;+\; m \frac{E_D}{B} C_P \;+\; (p+n_w)^2 \;+\; |\mathbf{T}| \frac{E_D^{(v)}}{B} C_P
\Bigg)
\end{equation*}
\section{Additional Experiments}\label{app_3}
\begin{figure}[ht]
\centering
\includegraphics[width=0.9\textwidth]{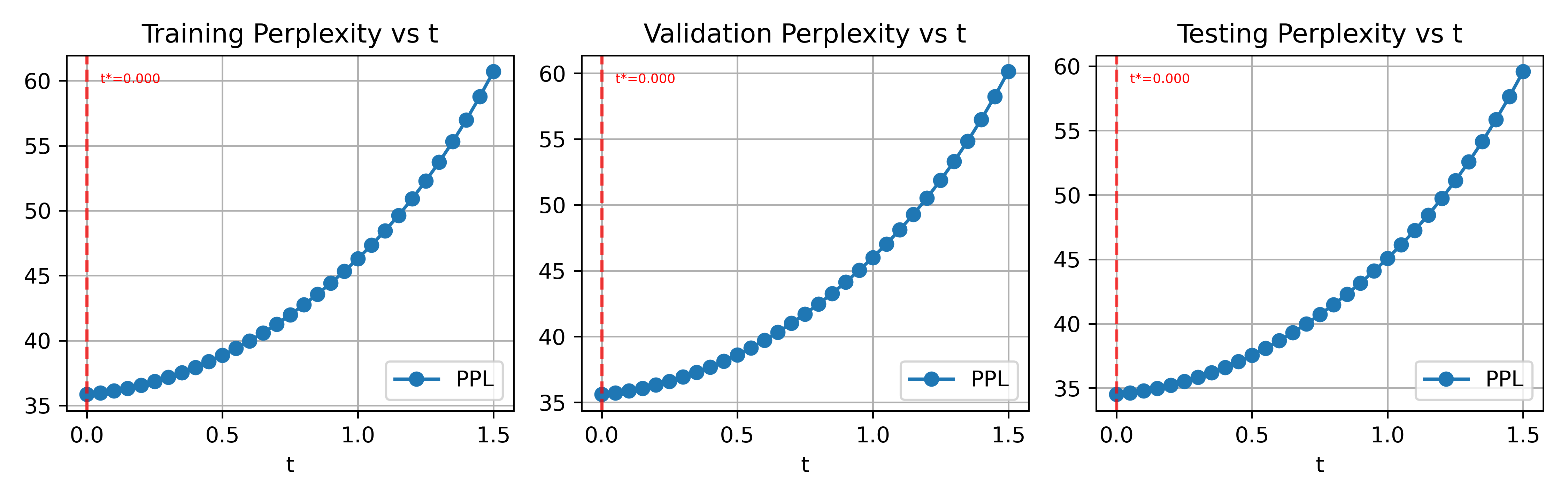}
\caption{HLS behavior (\(t^\ast=0\)) with \(\mathbf{N}_{WE}=0\).}
\label{fig:add_hls_0}
\end{figure}

\begin{figure}[H]
\centering
\includegraphics[width=0.9\textwidth]{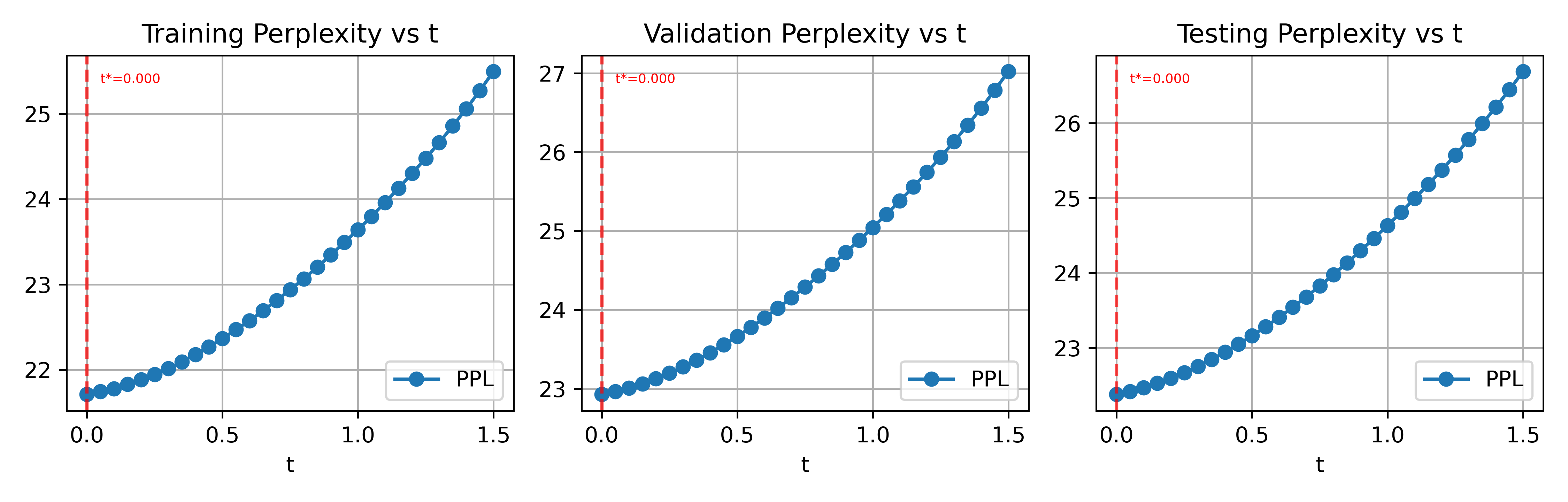}
\caption{HLS behavior (\(t^\ast=0\)) with \(\mathbf{N}_{WE}=1\).}
\label{fig:add_hls_1}
\end{figure}

\begin{figure}[ht]
\centering
\includegraphics[width=0.65\linewidth]{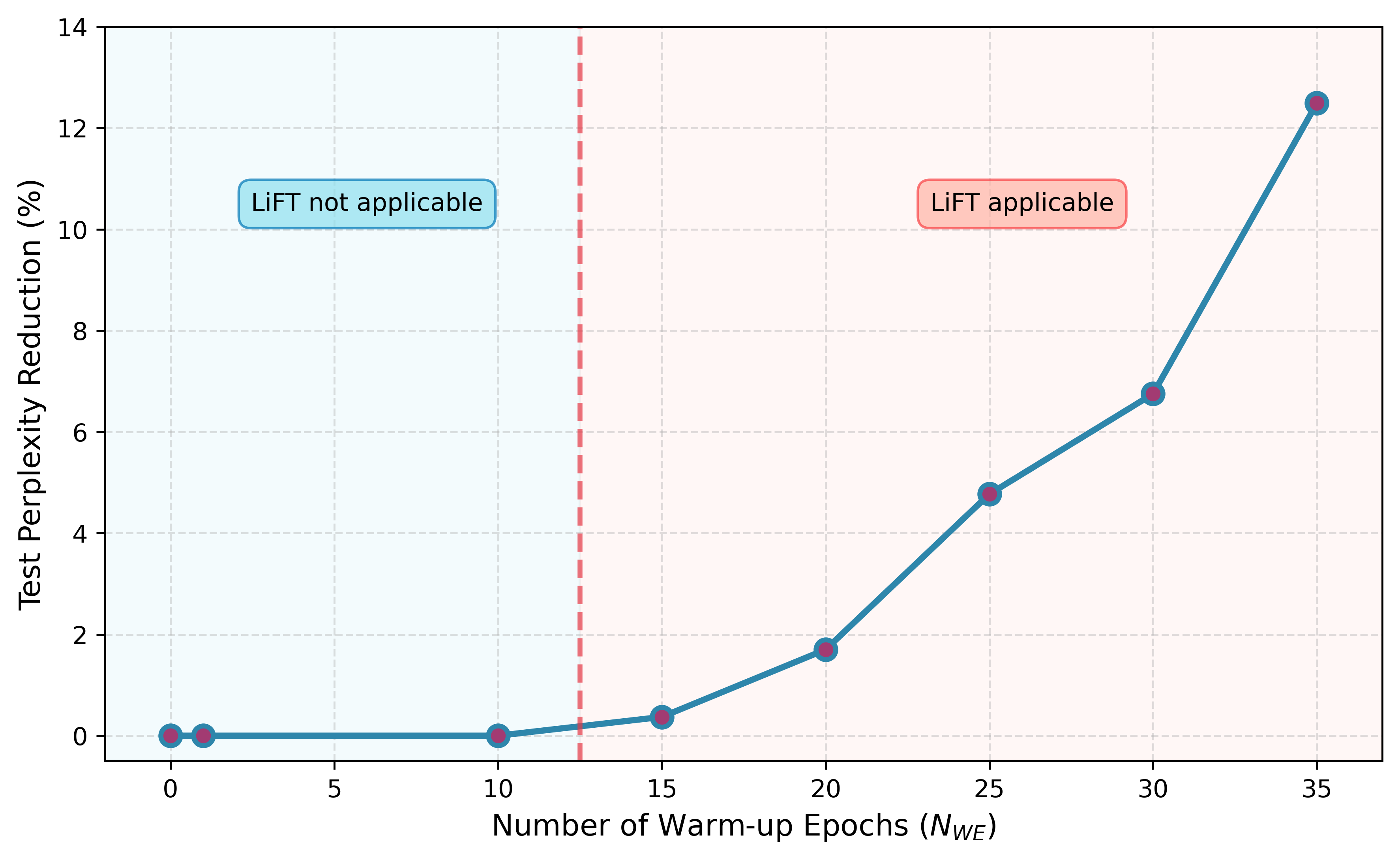}
\caption{Effect of warm-up training on test perplexity reduction.}
\label{fig:ppl_red}
\end{figure}

Table~\ref{tab:add_lift} and Figures~\ref{fig:add_hls_0} and~\ref{fig:add_hls_1} report the results obtained with no warm-up applied to the training data (\(\mathbf{N}_{WE}=0\)) and with a single warm-up epoch (\(\mathbf{N}_{WE}=1\)).
\begin{table}[h!]
\centering
\scriptsize
\caption{Model performance in the absence of overfitting.}
\label{tab:add_lift}
\resizebox{\columnwidth}{!}{%
\begin{tabular}{llllllcc}
\hline
\multicolumn{4}{c}{\textbf{HP}} & \textbf{\shortstack{Sample \\ Size}} & \textbf{\shortstack{Perplexity/ \\ Loss}} & \textbf{\shortstack{Without \\ HLS}} & \textbf{\shortstack{With \\ HLS}} \\
\cline{1-4}
\(\mathbf{N}_{WE}\) & \(\mathbf{N}_{TB}\) & \(\mathbf{N}_{R}\) & \(\mathbf{N}_{\text{FULL/MLP}}\)
 & & & & \\
\hline
  &  &  &  &  4718 & Training   & 35.84/3.58  & 35.84/3.58 \\
 0 & 3 & 2 & 1/2 &  487  & Validation & 35.60/3.57  & 35.60/3.57 \\
 & & & &  558  & Testing    & 34.51/3.54  &  34.51/3.54 \\
\hline
  &  &  &  &  4718 & Training   &  21.71/3.08 & 21.71/3.08 \\
 1 & 3 & 2 & 1/2 &  487  & Validation & 22.93/3.13  & 22.93/3.13 \\
 & & & &  558  & Testing    & 22.38/3.11  &  22.38/3.11 \\
\hline
\end{tabular}%
}
\end{table}
The first setting corresponds to the original GPT-2 model, while in the second setting the model parameters are updated by training for one epoch using the specified configurations. In both cases, we observe identical performance with and without HLS for the corresponding experiments. Notably, when a single epoch of training is performed, there is a substantial reduction in training, validation, and test perplexities. This indicates that limited fine-tuning can effectively adapt the model to a given dataset. However, as training continues for additional epochs, the training perplexity continues to decrease while validation and test perplexities degrade, signaling overfitting to the training data as shown the main results. These experiments clearly demonstrate that HLS, and consequently LiFT, is activated only when overfitting is detected.
When no overfitting is present, the method correctly recommends no parameter updates (\(t^\ast = 0\)), as illustrated in Figures~\ref{fig:add_hls_0} and~\ref{fig:add_hls_1}. In such scenarios, a simple fine-tuning exercise on the target data may be sufficient to improve test performance, as observed in the experiment with \(\mathbf{N}_{WE}=1\), where improvements in perplexity are obtained on the training, validation, and test datasets. Figure~\ref{fig:ppl_red} further illustrates the applicability of LiFT in the presence of overfitting, which typically arises as the number of warm-up epochs increases.

\end{document}